\documentclass[11pt,a4paper]{article}
\usepackage[hyperref]{emnlp2020}
\usepackage{times}
\usepackage{latexsym}

\usepackage{microtype}

\usepackage[utf8]{inputenc}
\usepackage[T1]{fontenc}
\usepackage{amsfonts}

\usepackage{microtype}
\usepackage{graphicx}
\usepackage{subfigure}
\usepackage{booktabs}
\usepackage{enumitem}
\usepackage{amsmath}
\usepackage{url}
\usepackage{tabularx}

\usepackage{verbatim} 

\usepackage{multirow}
\usepackage{color}

\definecolor{ourlightblue}{HTML}{E0ECF7}
\definecolor{ourdarkblue}{HTML}{092E6B}
\definecolor{msgrblue}{HTML}{4889f4}
\definecolor{msgrgray}{HTML}{e1e1e7}

\definecolor{botc}{rgb}{0.458, 0.488, 0.978}

\definecolor{humanc}{rgb}{0.8, 0.8, 0.8}

\definecolor{light-gray}{gray}{0.90}
\definecolor{dark-gray}{gray}{0.30}

\definecolor{aszlam}{rgb}{1.0, 0.0, .6}

\aclfinalcopy

\title{Deploying Lifelong Open-Domain Dialogue Learning }

\author{Kurt Shuster$\thanks{* Joint first authors.}$ \quad Jack Urbanek$^*$  \quad Emily Dinan \quad Arthur Szlam \quad Jason Weston\\
\\
  Facebook AI Research
}

\begin{document}

\maketitle

\begin{abstract}
Much of NLP research has focused on crowdsourced static datasets and
the supervised learning paradigm of training once and then evaluating test performance. 
As argued in \citet{de2020towards}, crowdsourced data has the issues 
of lack of naturalness and relevance to real-world use cases, while the static dataset
paradigm does not allow for a model to learn from its experiences of using language \cite{silver2013lifelong}.
%
%
 In contrast, one might hope for machine learning systems that become more useful as they interact with people.
In this work, we build and deploy a 
role-playing game, whereby human players converse with learning agents situated in an open-domain fantasy world.
We show that by training models on the conversations they have with humans in  the game the models progressively improve, as measured by automatic metrics and online engagement scores.
This learning is shown to be more efficient than crowdsourced data when applied to conversations with real users, as well as being far cheaper to collect. 
\end{abstract}

\section{Introduction}

Humans learn to use language 
over the course of their lives from the interactions they have with the world and other people.
Yet, the prevailing dominant paradigm in natural language processing (NLP) research is to build a fixed dataset from which to train a model and then freeze it, without any ability for the model to interact with humans using language at training time at all. While we need such  interaction in order to study human-machine communication to its full extent, constraints usually inhibit such research. 
Firstly, conducting such experiments can be costly. 
Many datasets in NLP are collected with crowdsourcing, whereby one pays the crowdworkers to perform interaction and annotation tasks. This leads to several issues, not least that 
research budgets for paying crowdworkers mean that data will have a limit.
Secondly, as crowdworkers are motivated by pay, not by interest in the actual tasks themselves, the data distribution may not match the desired one \cite{de2020towards}.

\begin{figure*}[ht!]
\includegraphics[width=16cm]{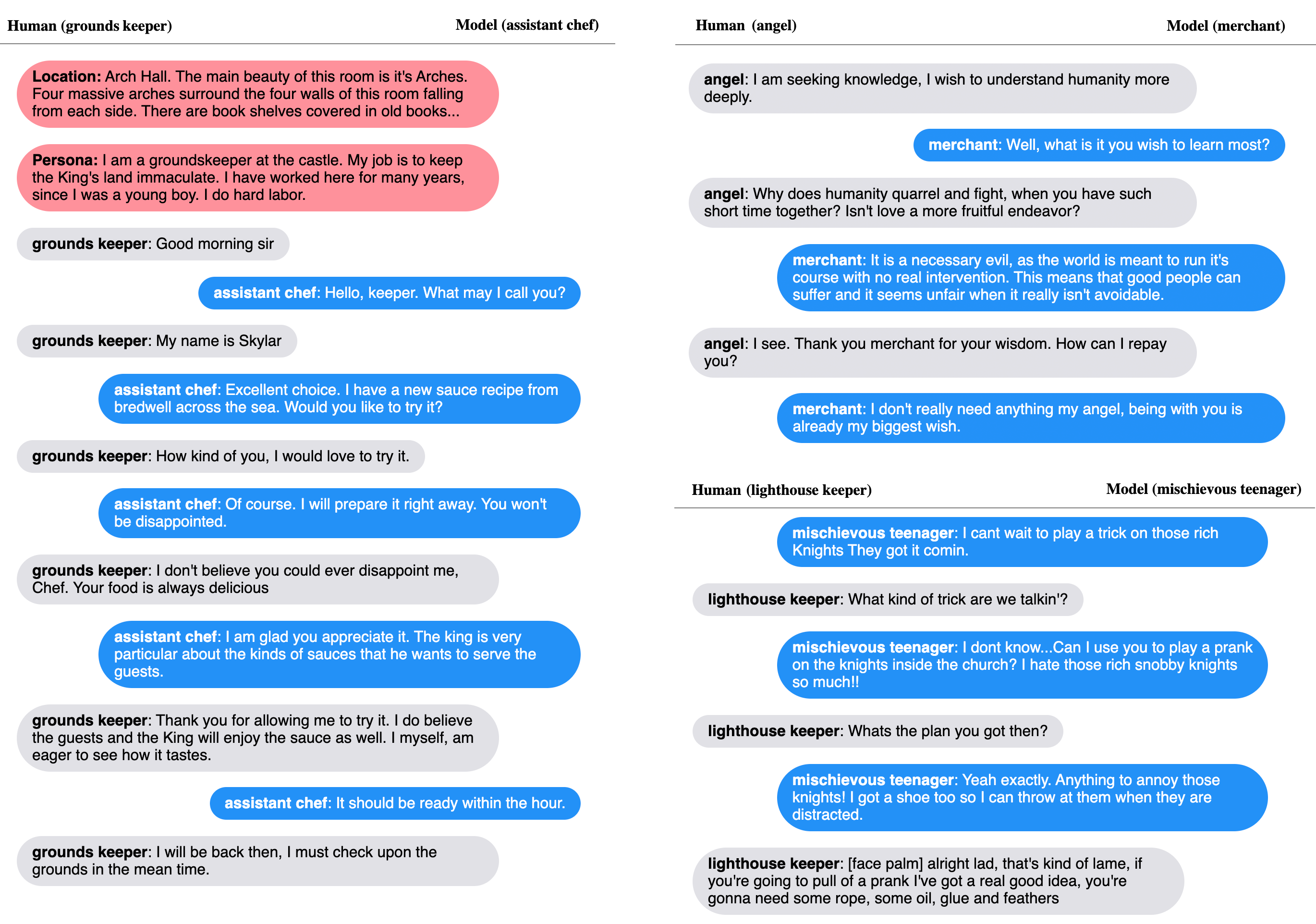}
\caption{Example collected dialogues from humans role-playing in our deployed system, conversing with models. (Left) a conversation complete with described location and player persona; (Right) excerpts from two other conversations (out of 41,131 collected) to demonstrate the diversity of the open-domain task.
\label{fig:cherries}
\label{ex:game-setup}
}
\end{figure*}

In this work we study the ability of an open-domain\footnote{In this work we study dialogue that can be about any topic but within the scope of a fantasy game world. Note this differs from open-domain dialogue talking about our world, e.g. the game players can talk about the sauce recipe from Bredwell across the sea (see Fig. \ref{fig:cherries}), but not about the pizza in Chicago.} dialogue model to iteratively learn from conversations with intrinsically motivated  humans. In order to engage humans at scale, we build and deploy a (free to play) game with a purpose \cite{von2006games} whereby human players role-play characters and converse with other characters (that are our learning models) situated within the game world.
We choose a fantasy game world, in order to maximize engagement. 
Our system iterates between collecting data of human-model interactions, retraining updated models on the newly collected data, and redeploying them. 
Simultaneously, it provides a natural metric to evaluate and compare models online using the continuation  rate of players (how long they continue playing).

We show that we can successfully collect, retrain and redeploy models that improve both offline automatic metrics and human continue rates. Our overall system is engaging enough that we can collect data 
at a rate that is $1/5^{th}$ of the price per utterance of crowdsourcing, where the cost of our method is the cost of advertisements that make players aware of the game. Moreover, the data we collect is also more effective per utterance at improving continue rates due to being more on-distribution than crowdsourced data.
As our models improve, these rates improve as well, as the continuation rate increases -- meaning relatively more data is collected.
Overall, our work provides good evidence that lifelong dialogue learning in deployed systems with intrinsically motivated humans (rather than crowdworkers) can be successful,  in particular by embedding such learning within games.


The training code and parameters of the models deployed, and the data collected in this work will be made publicly available for reproducibility and further research by the community\footnote{Available at: 
\href{http://google.com}{\nolinkurl{parl.ai/projects/light}}}.



 

\section{Related Work}

\paragraph{Open-Domain Dialogue}

Dialogue in the open-domain setting, wherein the conversation involves chat about any topic, rather than a specific goal-directed topic,  is commonly studied in the train/valid/test static dataset paradigm utilizing supervised learning.
A number of crowdsourced or scraped datasets have been developed to that end, including  Daily Dialogue \cite{li2017dailydialog}, PersonaChat \cite{li2016persona}, Empathetic Dialogues \cite{rashkin2019empathetic} and Wizard of Wikipedia \cite{dinan2018wizard}; see \citet{huang2020challenges} for a review.

\paragraph{LIGHT}
In this work we specifically focus on the open-domain dialogue setting of LIGHT \cite{urbanek2019learning}. LIGHT focuses on situated characters playing character roles that can talk about any topic, within the context of a medieval fantasy world. This setting is known to be engaging for human role-players,  and also alleviates some safety concerns in that the role-playing means they should not divulge personally identifying information.
The authors crowdsourced a dialogue dataset consisting of 8.5k episodes and 111k utterances, which they publicly released. We refer to this as LIGHT MTurk data, or LIGHT data for short, in the rest of this paper. In this work we utilize this data to build a deployed system whereby players can converse with models, and we can study lifelong learning with these models using the information in these new conversations.

\paragraph{Lifelong Learning}

Lifelong learning is a machine learning paradigm whereby deployed models can interact
with the world and iteratively improve themselves from the things they learn,
eschewing the standard approach of a fixed  training set from which a model is trained once \cite{silver2013lifelong}. We note there are other closely related concepts to the topics in this work, such as incremental learning \cite{castro2018end}, 
continual reinforcement learning \cite{ring1994continual} and
 never-ending learning  \cite{carlson2010toward,mitchell2018never}.
 
\paragraph{Continual Dialogue Learning}

Learning from dialogue interaction is  common in reinforcement learning settings, where the feedback is a scalar rather than solely the dialogue messages themselves 
\cite{levin2000stochastic,schatzmann2006survey,rieser2011reinforcement,liu2017end,serban2017deep}, which is most common in a goal-oriented setting where 
completion of the goal can provide such rewards.  
In this work we study learning from open-domain interactive dialogue messages, not from rewards. 

Closer to our work, is the self-feeding chatbot \cite{hancock2019selffeeding},
whereby it is shown that models can be used to collect data to improve themselves via crowdsourcing utilizing the PersonaChat task.
Related approaches have also been applied to the more limited case of question answering
\cite{li2016dialogue,li2016learning}, or in simulation \cite{mazumder2019lifelong} as well.
\citet{liu2018dialogue} applied such an approach to goal-oriented tasks.
Our work differs from these works
in that we study a deployed user-facing system in a rich open-domain environment, 
rather than more limited data from paid crowdworkers, and thus
study a more realistic setting.

\paragraph{Deployed Dialogue Systems}

While there are a number of deployed open-domain virtual assistants, many of these products are not ideal platforms for the research community. Their proprietary nature and commercial importance, coupled with privacy concerns, means they are neither accessible to researchers, 
nor amenable to public reproducible research.  
A near-exception is the Alexa challenge \cite{ram2018conversational} which allows university-based researchers access to a commercial user-base for the span of the competition, 
however, the data and models are also not released to the rest of the research community.

\section{Open-domain dialogue as a game}\label{sec:ldrpg}

In this section we describe the game that we will build
and deploy,  which is a dialogue role-playing game. 
It is a game with a purpose, and as such is
designed to both train and evaluate open-domain dialogue agents.

\paragraph{Core Game}
The core game involves pairing two agents in a given setting -- where one is a human player and the other is a dialogue agent with an underlying machine learning model.
The two players are assigned characters, with given names and backstories (personas),
and their current location and its description. See Figure \ref{ex:game-setup} 
for examples. Each player's goal is simply to act out (role-play) their character's dialogue in the given situation.  We refer to one such dialogue episode as a {\em mini-game}. Dialogue in the game is in English.

\paragraph{Role-Playing (Acting) Score}
We take advantage that role-playing is a pursuit that a large number of human players find
fun \cite{horsfall2011study}, and are hence naturally engaged in the open-ended nature of 
this process. However, to encourage and further motivate players to play their characters well,
we introduce the concept of an (automated) dungeon master (DM), who will assess the quality of the player's role-playing. For each dialogue turn, we apply a learned model to the human player's dialogue, which assesses how likely their utterance is given the context. We convert this to a score, between 1 and 5 stars, that is presented to the human player, to reward them for good acting. While this signal is noisy, because our DM model is not perfect, it gives motivating feedback to the players to continue playing.

\paragraph{Other Gamification Steps}
The acting scores (between 1-5 stars per turn) are accumulated, and a player's total score is presented on a leaderboard compared to all other players, providing further motivation to reach the top of the leaderboard. 
We also award ``badges'' if, for a given dialogue, a certain number of points are collected (11 for 1 badge, 16 for two):
the badges represent the characters in the game, motivating the desire to role-play all the characters in the game, and collect all the badges.

\paragraph{Game Loop}
Each dialogue (mini-game) consists of 6 turns of dialogue per agent (12 total). At the end of the mini-game the human player is presented with four choices: (i) choose to move to a new location, where they will continue to play this character, but meet a new character to converse with; 
(ii) stay in the same room but wait for a new character to arrive to converse with; (iii) change to role-play a completely new pair of characters in a new setting; or (iv) end the game.
These choices encourage the player to choose another mini-game that they are most interested in, and the variety of mini-games gives different role-playing possibilities, making the dialogue data more diverse.

\paragraph{License Agreement and Public Release}
Upon entry to the game, players are asked to agree to the use and release of the resulting game data as a publicly available dataset for research purposes. They are urged to stick to their assigned characters in the game, and hence should not use any personally identifying information, which the terms also tell them explicitly not to share. In the released data, no other information about the player is retained except for the messages they send in the game.

\paragraph{Game Safety}
We employ a safety classifier \cite{dinan2019safety} on both human and model turns. For safety reasons, we limit our dialogue models to be retrieval models, so that we could vet the entire set of candidates  for offensive language before run-time. The set
of settings and character personas were all also vetted for offensive language.
Additionally, gender bias concerns have been previously studied within the available LIGHT MTurk training set \cite{dinan2019queens}, and we make use of that publicly available data here as well.
We note that, compared to other deployed dialogue systems, there is an extra level of indirection due to playing characters in a game that makes language relatively less offensive. For example, a thief in the game saying ``I'm going to steal your money'' to another game character is far less offensive compared to a digital assistant saying it directly to a human user.

\section{Lifelong Dialogue Learning}

\subsection{Models}\label{sec:models}

\paragraph{Retrieval Models}
All the models we have currently deployed are retrieval models (see previous discussion of safety). In particular, we use the Poly-Encoder (PE) Transformer architecture as a base \cite{humeau2019polyencoder}, as it provides state of the art results compared to other retrieval models, whilst being tractable to deploy. 
PE encodes the context with a standard bidirectional transformer, but produces an encoding into a fixed small number of codes, $N$. We tried values of $N=5$ and $N=20$. 
Each label candidate then attends to these codes before producing a final matching score. The model is trained with cross-entropy given the correct label, and by subsampling negative examples from the given training batch.

\paragraph{Architecture and Training Choices}
We employ the 90M and 622M parameter models from
\cite{roller2020recipes} that have been pre-trained on 1.5B training
examples from pushshift.io Reddit, which we then fine-tune. We also consider two other enhancements, chosen to mitigate problems that we observed with the models:
(i) negative context training, whereby negatives are also selected from the immediate dialogue history as well as the batch
which can help reduce a model's tendency to repeat itself \cite{holtzman2019curious,welleck2019neuraltext};
and (ii) decoding control \cite{see2019goodconversation} whereby at decoding time responses are rescaled before scoring based on their specificity (normalized inverse document frequency). The latter can control the genericness of the responses, which is known to affect human judgments.

\paragraph{Generative Models}
In addition to the deployed models, we also perform training and automatic evaluation metrics on generative models offline, where safety concerns are less important as the models are not user-facing. 
We employ an encoder-decoder Transformer architecture using the state of the art pre-trained 2.7 billion parameter BlenderBot model \cite{roller2020recipes}, which we fine-tune on our task. 

\paragraph{Agent Dialogue Model}
Training a dialogue model involves one of the setups described above, and a set of (dialogue context, correct label) pairs. We will train on such pairs both from crowdsourced data and data collected within game in our lifelong learning setup.
All training and experiments are performed using the ParlAI software framework \cite{miller2017parlai}.

\paragraph{Acting Score Model} We can apply a retrieval model to also score the human's role-playing abilities. In this case, the context is the entire dialogue history, setting and the player's character persona as input to the encoder, while the candidates to score are the ones from the training set, as usual, plus additionally the human's (player's) actual response.  For speed, the encoder can actually be run while the human player is typing, as it does not depend on their response, which is treated as a candidate label instead. The score given to the user is then proportional to the human response's rank amongst all the candidates\footnote{The player is awarded 2 stars if their response is in the top 2000, 3 stars in the top 1000, and 4 stars in the top 100.}.

\subsection{Iterative Data Collection and Training}

After collecting a certain amount of episodes of conversational data between humans and models, 
one can consider using this data for training. 
We utilize the following observation: while the model utterances may contain many mistakes, it is assumed that a human sufficiently  engaged provides high quality responses, even if the model responses are mistakes, and can thus be treated as gold, and used as a fully supervised signal. We thus separate the dialogue data into all possible (context, next utterance) pairs, and then only consider  the pairs with human next utterances as training data. We also compare this to further filtering this set by scoring the quality of the human utterances, discarding those episodes (mini-games) with lower quality. We use the acting score model previously described for this purpose, summing the scores obtained across an episode, and discarding the episode if this value is less than $C$, 
where $C$ is a hyperparameter tuned on the validation set.

After training our model from a given round of collection, we can go back to the collection process utilizing instead the new model that has been trained on more data. The hypothesis is that the
higher quality the model is: (i) the higher quality the human data will be as well; and (ii) the more likely the human players are to converse longer, increasing the data set size by larger amounts. 

\subsection{Deployment-based Evaluation}
\label{sec:deploy_multi}

Apart from the collection-training cycle of our deployed lifelong learning setup,  one can also 
in parallel perform evaluation.
For each separate mini-game (episode of dialogue) we can potentially deploy a {\em different} model for human-model conversation.
We maintain a pool of models with differing architectures or hyperparameters, and select randomly from the pool in each episode. For any given episode we record whether the player continued playing to the next mini-game or not, which we refer to as the continue rate. We can measure the quality of a model using its averaged continue rate over all players and episodes. In this way we can also perform model selection online. 


\begin{table*}[h]
\center
\begin{small}
\begin{tabular}{lccccc}
 \toprule
 Data Type  & Num. Epsiodes & Num. Utterances & Num. Human Utterances & Unique Locations &  Unique Characters \\
 \midrule
 Training   & 41,131        & 461, 984  & 230,992      & 587  & 630 \\
 Validation &  500        &  5,936    & 2,968      & 231 & 463\\
 Test       &  1000       &  11,822 & 5,911        & 296 & 569  \\
\bottomrule  
\end{tabular}
\end{small}
\caption{
Data statistics of our lifelong learning deployment at the point where we froze collection for experiments reported within the paper and subsequent data release.
\label{tab:dataset}
}
\end{table*}

\begin{table*}[h]
\center
\begin{small}
\begin{tabular}{lcccccc}
\toprule
         & Num. &	Num.    &	Num. Human	        &  Unique  &	Avg. Human	& Number of \\
Dataset	 & Episodes &	Utterances &	 Utterances	 & Tokens &	Utt. Length	& Humans \\
\toprule
PersonaChat \tiny\cite{zhang2018personalizing} & 8,939 & 131,438 & 131,438 & 18,688 & 11.9 & \tiny{UNKNOWN} \\
Wiz. of Wikipedia \tiny\cite{dinan2018wizard} & 18,430 & 166,787 & 166,787 &52,490  & 19.7 & \tiny{UNKNOWN} \\
Empathetic Dialog \tiny\cite{rashkin2019empathetic} &  24,850 & 64,636 & 64,636 &  19,458 &  15.3 & 810\\
Daily Dialog \tiny\cite{li2017dailydialog}            & 22,236 & 87,170 & 87,170 & 20,673 & 14.5 & \tiny{UNKNOWN} \\
LIGHT MTurk \tiny{\cite{urbanek2019learning}}	& 8,538 &	110,877	& 110,877	& 33,789 &	18.3 & 1,052 \\	
LIGHT WILD \small{\em (this paper)} &	41,131 &	461,984 &	230,992	& 47,526	& 11.9
&	13,188 \\
\bottomrule  
\end{tabular}
\end{small}
\caption{
Comparison of statistics of the open-domain dialogue data collected during our lifelong learning deployment (bottom row) compared to several existing (mostly crowdsourced) datasets.
Our data is around twice as large in terms of  human utterances than these datasets, and 4x as large in terms of dialogue utterances (as our data consists of human-model conversations), while the cost to collect our data was only ${1/5}^{th}$  of the price per utterance of LIGHT MTurk, see Sec. \ref{sec:cost_dataset}.
\label{tab:dataset_compare}
}
\end{table*}

\begin{table*}[h]
\center
\begin{small}
\begin{tabular}{lccc|ccc}
\toprule
 & 	\multicolumn{3}{c|}{Retrieval Model (Hits@1/20 $\Uparrow$)}		&	
\multicolumn{3}{c}{Generative Model (PPL $\Downarrow$)} \\		
Model  & LIGHT Test &	LIGHT Test Unseen	& WILD Test &	
LIGHT Test &	LIGHT Test Unseen	& WILD Test \\ 	
\toprule
Round 1 & 87.12 & 82.43 & 81.61 & 12.67 & 11.81 & 13.42\\ 		
Round 2 & 87.65 & 82.70 & 84.60	& 12.57 & 11.74 & 12.31 \\	
Round 3 & 87.72 & 83.48 & 87.63	& 12.54 & 11.75 & 11.79 \\	
\bottomrule
\end{tabular}
\end{small}
\caption{Three rounds of training in our lifelong open-domain dialogue learning setup. Both retrieval and generative models trained on the data from the three rounds improve across both metrics on all three test sets.
\label{tab:3rounds}
}
\end{table*}

\begin{figure*}[h]
\includegraphics[width=8cm]{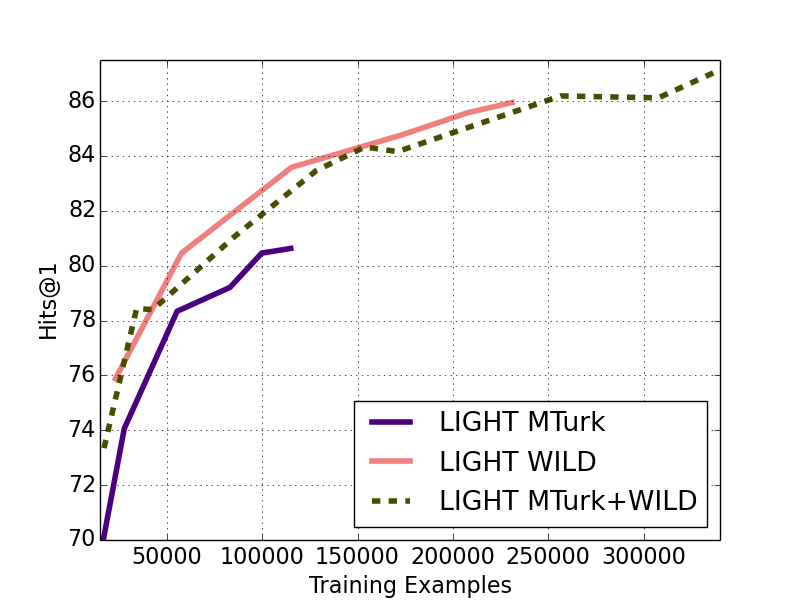}
\includegraphics[width=8cm]{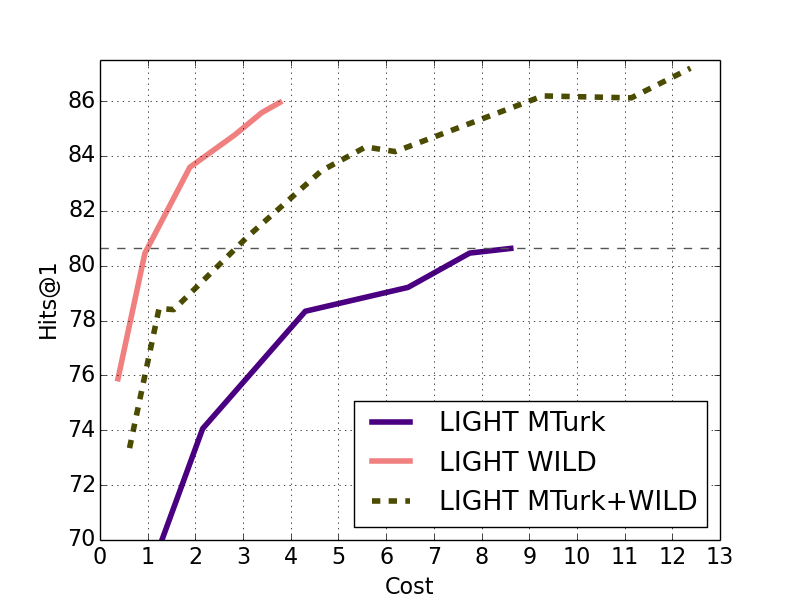}
\caption{Hits@1/20 Accuracy on the LIGHT WILD validation set as a function of the number of training examples (left) or the cost of data collection (right). The cost axis is in units scaled by the cost of LIGHT WILD collection required to achieve the same performance as using the entire LIGHT MTurk dataset; it is more than 8$\times$ cheaper to use LIGHT WILD examples than LIGHT MTurk examples to achieve an accuracy of  80.63\%. We also show performance for models which equally sample data from LIGHT MTurk+WILD datasets for training; utilizing all the data from both sources yields the best performance. However, LIGHT WILD data gives better accuracy improvements per training example (left plot).
\label{fig:learningcurve}
}
\end{figure*}

\begin{figure*}[h]
\includegraphics[width=8cm]{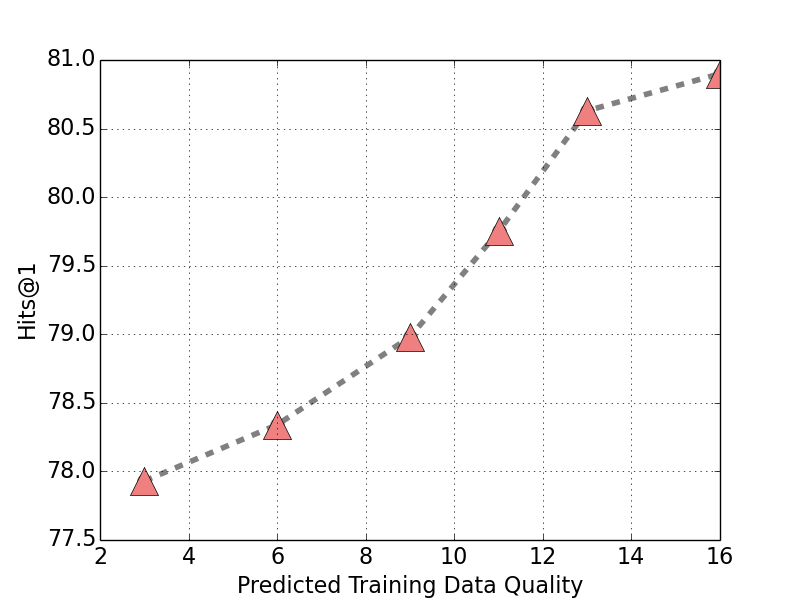}
\includegraphics[width=8cm]{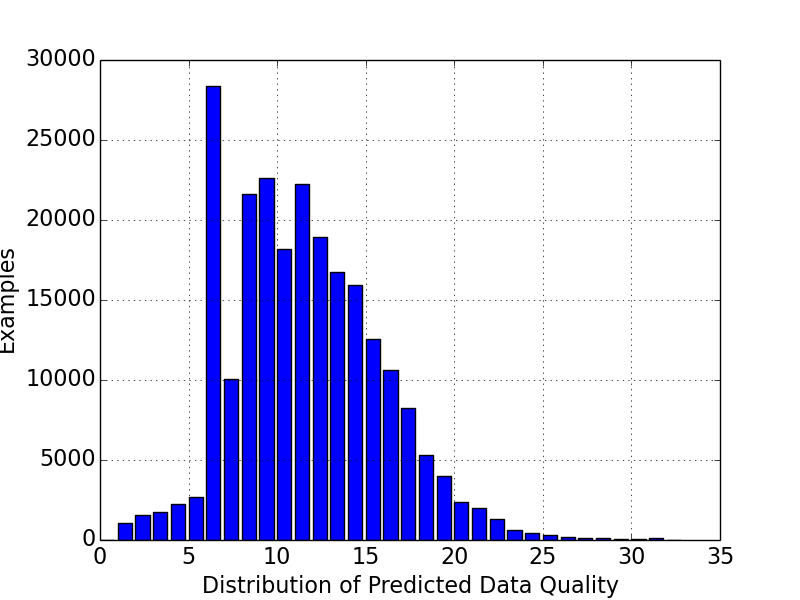}
\caption{{\bf Predicted Data Quality}. Left: Hits@1/20 accuracy on the WILD validation set when training with LIGHT MTurk + 10,000 examples from the WILD training set of a given predicted quality level, see Sec. \ref{sec:exp_quality}.  Data that is predicted to be higher quality yields improved validation accuracies.
Right: The distribution of data quality predictions over the training set. A spike is seen at quality bin 6 because that is the lowest score one can achieve when completing a full episode (1 star per turn is awarded at minimum). Values lower than bin 5 indicate incomplete low-scoring  episodes.
\label{fig:data_quality}
}
\end{figure*}

\begin{table*}[h]
\center
\begin{small}
\begin{tabular}{lcccccc}
\toprule
Model  &  Train Data & Negative Context & Decoding Control &	Continuation Rate \\ 
\toprule
90M PE 	&  LIGHT 	     & yes & no & $72.2 \pm 1.9\%$\\	
90M PE 	&  LIGHT 	     & yes & yes & $74.1 \pm 2.0\%$\\
90M PE 	&  LIGHT + WILD  & yes & no & $73.6 \pm 1.8\%$\\
90M PE 	&  LIGHT + WILD  & yes & yes & $75.2 \pm 2.0\%$\\	
\midrule
622M PE &  LIGHT 	     & no & no & $68.2 \pm 1.4\%$\\	
622M PE &  LIGHT 	     & yes & no & $69.9 \pm 1.9\%$\\	
622M PE &  LIGHT 	     & yes & yes & $69.9 \pm 2.0\%$\\			
622M PE &  LIGHT + WILD  & yes & no & $70.6 \pm 2.1\%$\\					
622M PE &  LIGHT + WILD  & yes & yes & $71.8 \pm 1.9\%$\\
\bottomrule
\end{tabular}
\end{small}
\caption{Deployment-based Evaluation, comparing several metrics on data collected during Round 2 of collection. 
\label{tab:continuerates}
}
\end{table*}

\begin{table}[h]
\center
\begin{tabular}{lc}
\toprule
Model variation &    {\small $\Delta$ Continue Rate} \\ 
\toprule
+ WILD train data (Round 2)  & $+1.3 \pm 0.7 \%$ \\
90M $\rightarrow$ 622M parameters PE & $-3.2 \pm 0.7 \%$ \\
+ Negative context training & $+2.6 \pm 1.3\% $\\
+ Decoding control               &  $+2.5 \pm 1.1\%$ \\
\bottomrule
\end{tabular}
\caption{Deployment-based Evaluation:
changes in continue rates for various model variants.
\label{tab:continuerates2}
}
\end{table}

\section{Experiments}

\subsection{Rounds of Learning}
We performed three rounds of our lifelong learning setup.

\paragraph{Round 1} consists of models trained on LIGHT MTurk data only. We train the retrieval model variants described in Section \ref{sec:models}, and deploy them within the game.
\paragraph{Round 2} consists of models trained on LIGHT MTurk data + 50,982  WILD examples collected from the deployment of the Round 1 models, and again deploy these within the game.
\paragraph{Round 3} consists of models trained on LIGHT MTurk data + 50,982 examples from Round 1 deployment + an additional 180,010 examples collected from Round 2 deployment.

\subsection{Data Collection}

While our setup is a lifelong learning setup and the models are still currently deployed and collecting data, for this paper we froze the collection at a given point in order to provide a data release and provide experimental results. The data statistics for the total newly collected dataset, called LIGHT WILD, over all rounds is shown in Table \ref{tab:dataset}. Validation and test sets were extracted from a portion of the data\footnote{For validation and test we only use complete conversations, and where the player scored $\geq$ 9  stars, to build higher quality evaluation sets.} from Round 2. 

Table \ref{tab:dataset_compare} compares this dataset to several existing commonly used open-domain dialogue datasets. The number of episodes and dialogue utterances are larger than many existing datasets, e.g. four times as many as LIGHT MTurk, and almost eight times that of Empathetic Dialogues.
Uniquely, our dataset contains human-model conversations, hence the total number of human utterances is actually half of the utterances, which is still twice as large as the number in LIGHT MTurk.
Our dataset also has a large degree of diversity, which is important for tasks in general, and especially for 
open-domain dialogue. The number of unique locations and  roles that can be played by speakers (characters) 
is large (587 and 630, respectively). The number of players of the game at the time of freezing was over 13,000, which also makes the diversity far larger than typical crowdsourced datasets, e.g. LIGHT MTurk involved 1,052 crowdworkers and Empathetic Dialog involved 810 crowdworkers. Finally, the number of unique tokens is larger in LIGHT WILD, indicating the diversity of language used.

\subsection{Analysis of Results}

\subsubsection{ Performance by Round}

While we only deployed retrieval models,
we report experiments training both retrieval models and generative models on the data from the three rounds, selecting best hyperparameters using the validation set. We report the performance on three different test sets: LIGHT (MTurk) Seen and Unseen test sets \cite{urbanek2019learning}, where unseen means that the test locations do not overlap with the training set locations, 
and our WILD test set. The results are given in  Table \ref{tab:3rounds}. They show a steady increase in the Hits@1/20 metric (Top 1 accuracy given 19 random distractors) for the retrieval models over the rounds on all three test sets, and a similar decrease in perplexity (PPL) for the generative models. 
In particular there is a large jump in the performance on the WILD Test set between Rounds 1 and 2 as the training set switches from crowdsourced to in-distribution WILD data, and a further increase in Round 3 as more data is again collected and retrained on. While our WILD data is of a different distribution to the two LIGHT (MTurk) test sets, the data collection from our lifelong learning setup still gives gains on those tests as well. Our reported numbers,  as far as we are aware,  are the best reported numbers on these datasets, e.g. the original LIGHT paper reports 76.5\% and 70.5\% for the Seen and Unseen test sets, respectively (compared to our 87.72\% and 83.48\%). Overall, we see clear gains from the extra data collected in our setup.
\subsubsection{Lifelong Learning Curves}

We construct learning curves given all the collected data to analyze the performance gain per new training example. We plot Hits@1/20 accuracy on the WILD validation set 
against the number of training examples, 
comparing data from WILD collection to LIGHT (Mturk). We also consider a 50/50 mix, where we equally sample from the two sources LIGHT+WILD to provide the next training example. 

Figure \ref{fig:learningcurve} (left) shows the results. We observe that on a per-example
basis our WILD data gives more accuracy gains than LIGHT MTurk data, e.g. 83.59\%   for WILD compared to 80.63\% for LIGHT, when limiting WILD to the same training set size as the total size of LIGHT.
As the WILD dataset is more than twice as large this monotonically improves further, up to  85.95\% using all of the WILD data.  We observe that the improvements have not saturated and that further lifelong learning will bring further model improvements. 
Combining the two data sources, as shown in the LIGHT+WILD plot, brings yet further gains, up to 87.2\%.
Overall, our collected WILD data  has high quality as a learning signal for training models.

\subsubsection{Cost Learning Curves} \label{sec:cost_dataset}

We plot similar learning curves, but as a function of the cost to collect the data instead of the number of training examples instead, see Figure \ref{fig:learningcurve} (right).
Although we do not pay players to play the game, we did spend money to advertise the game online in order to attract players. We compare the cost per WILD example relatively to the cost per LIGHT (MTurk) example, where the x-axis is scaled in units that are multiples of the cost required to achieve  80.63\% using WILD data (as this is the performance of using all the LIGHT MTurk data together). We observe that it costs over 8x more to achieve the same accuracy using LIGHT (MTurk) data (see dashed horizontal line). The actual cost per utterance of WILD data is around 1/5$^{th}$ of the price of LIGHT MTurk data, but more than that, it is also relatively more effective per utterance in terms of metrics. For the same amount spent there is a large gap between the two systems, for example using all the WILD data gives a performance of 85.95\%, whereas for the same spend LIGHT MTurk only achieves $\sim$77.5\%.
Overall, WILD deployment is relatively a very cost effective strategy.

\subsubsection{Deployment-based Evaluation}

Our lifelong learning setup deploys multiple models (see Sec. \ref{sec:deploy_multi}) at the same time randomly assigning them to concurrent users per episode (mini-game). We can thus directly compare the quality of models 
 via their continue rate. 
 
 Continue rates during Round 2 
of collection comparing several model variants are given in Table \ref{tab:continuerates}.
Continue rates are in the range of 68\%-75\%, depending on the model, and we observe some clear trends.
Most importantly,
for both model sizes tried, LIGHT+WILD trained models are superior to LIGHT only trained models, showing 
that our deployment/train cycle produces better models as judged by humans.
Secondly, other factors in model design are important too, and our setup can effectively evaluate those. 
In particular, for both model sizes it was found that both our negative context training and decoding control enhancements (see Sec. \ref{sec:models}) improve the continue rate, with both methods used together improving more.
We confirm these conclusion in Table \ref{tab:continuerates2} where we show the change in continue rates when independently adjusting one of these factors, by averaging over model continue rates for other factors of variation. 

We also observe the unexpected result that the larger models perform worse than the small models across the board on continue rates. Deeper analysis given in appendix \ref{appendix:smallbig} suggests that while the larger model makes less mistakes, it is more often seen as boring, which would reasonably impact a player's desire to continue playing. Understanding and controlling this trade-off should be studied further.


\subsubsection{Data Quality} \label{sec:exp_quality}

Not every player is as engaged in the game as every other player, or produces as high quality dialogue. We hypothesize that we can predict which 
players produce higher quality data via the acting score model (Sec. \ref{sec:models}), and that such higher quality data is relatively better for training models. 

Figure \ref{fig:data_quality} (right) shows the distribution over the WILD training set of predicted quality using the acting score model. We observe 83.7\% of the episodes have a score above the minimum value of 6 (there are 6 turns, and on each turn a score between 1-4 is awarded, explaining the spike at the value of 6). Scores below 6 indicate incomplete dialogues, which only account for 4.0\% of the data.

To assess whether these scores are indeed indicators of data quality, we selected an equal amount of 10,000 examples from each of the bins (1-5), (6), (7), \dots, (16)  (grouping 1-5 together to make that group large enough) and compared them as training sources. We train a set of retrieval models on these training sources, where each model also has access
to all of the LIGHT MTurk data (111k examples) in addition to the WILD 10k from their respective bins. The results are given in Figure \ref{fig:data_quality} (left).
We observe a monotonic improvement on the WILD validation set with increasing predicted quality.
We see similar, but smaller gains on the LIGHT (Seen) validation set as well, e.g. 86.59\% for quality bin 6, and 87.11\% for quality bin 16.

While we can clearly select lower or higher quality data, we can also ask the question whether some of the data is so low quality we should simply remove it from the training data in order to get better performance. Experiments show that is not the case, and that even the lowest quality data does provide a useful signal, e.g. performance drops slightly from 87.06\% to 86.69\%  on the WILD validation set if we remove bins lower than 6, but otherwise training on all other data, and to 85.38\% if we remove bins lower than 9.


\subsubsection{Observations on Gamification}

Just as the design of a crowdsourcing task will affect the cost and quality of data, this is likely even more the case
in the design of a game. If the design is poor, players will not play it at all; whereas in contrast to paying crowdworkers, if players really like a game, they are willing to pay to play it.
Accordingly, the plots we presented in Figure \ref{fig:learningcurve} represent the results of our particular game design; there may well be a design with vastly more cost efficient learning rates. While a full of study of the elements of game design is outside of the scope of this paper, we note that for adjustments we did make to the game after initial deployment we observed large changes in user behavior. For example, after the addition of the three user controls for how to continue the game loop after an episode is finished (as described in Sec. \ref{sec:ldrpg}), compared to only a single choice, we saw an increase in the continue rate by $3.3 \pm 1.6 \%$ when using the same model. 

Model quality also affects cost and quality of the data collected. Noting the effects of changing gamification options (alongside other hard-to-track circumstances) we only report continue rate comparisons between models relative to runs in the same batch. Still, players' enjoyment of these models (as estimated by continue rate in Table \ref{tab:continuerates}) directly changes how much they engage with the game. As such it is more expensive to test models that are worse for the game experience (which we would consider fair from a player perspective). Hence, as models improve, costs actually go down, enabling to collect data at higher rate.

\subsubsection{Analysis of Data Distribution}

We can compare the dialogue data collected within our deployed system to crowdsourced data from LIGHT MTurk. 
We analyze over and underexpressed words in our dataset compared to the latter.

Calculating the top 70 most overexpressed words, all overexpressed at least 3.5x relative to crowdsourced data, we note several interesting observations about our data's distribution:\\
- { There are more natural endings to conversations}: e.g. ``goodbye'' ($4\times$) and ``bye'' ($4\times$) are overexpressed.\\
- { There are overexpressed words associated with aggression}:
  ``stab'' ($8.5\times$), ``dagger'' ($6.1\times$), ``club'' ($5.5\times$), ``kills'' ($4.9\times$),  ``blade'' ($4.2\times$).\\
- { There are overexpressed words associated with overtly friendly actions as well}: 
   ``smiles'' ($12.9\times$), ``nods'' ($10.9\times$),  ``kiss'' ($6.1\times$), ``hug'' ($3.7\times$), and ``bows'' ($5.9\times$).\\
- { There are more mentions of adventuring}: 
   ``quest'' ($5.4\times$), and other similar words not in the top 70 are overexpressed as well, such as ``adventure'' ($2.5\times$) and ``mission'' ($2.1\times$).\\
- { There is an increased use of slang}:  ``ur'' ($93\times$), ``u'' ($28\times$),  ``yo'' ($5\times$), ``dude'' ($6\times$). We note that some emojis exist in the dataset as well, which do not appear at all in the crowdsourced data.\\

In contrast, looking at the 70 most underexpressed words, all underexpressed by a factor of at least $1.3\times$, we observed the following patterns:
- { Less mentions of village and farm life}:
``peasant'', ``fields'' (both $2\times$ underexpressed), ``farm''  and ``crops'' (both $1.9\times$), ``harvest''  ($1.8\times$), ``villagers'' ($1.7\times$),  and ``work'' ($1.4\times$). \\
- { Less mention of passages of time}:
``week'' ($2.1\times$),  ``year'' ($1.9\times$),  ``days'' ($1.8\times$).

Overall, we see a pattern that game players seek more exciting conversations, 
involving more emotional, action-packed interactions such as seeking quests, 
whereas crowdworkers 
are more even-keeled, and able to discuss dry topics such as last year's harvest or taxes  with more frequency. 
This is not unexpected as game players often seek immediacy and 
larger-than-life experiences \cite{grodal2000video}.

\section{Conclusion and Future Work}

We have presented a fully realized system for improving 
upon an open-domain dialogue task
by utilizing a deployed game for lifelong learning.  
Detailed experiments showed that the one can collect high quality data that improves both automatic offline metrics and user engagement metrics when used for training models.
We find this exciting because this approach shows it is possible to build continually improving models that learn from interacting with humans in the wild (as opposed to experiments with paid crowdworkers), which represents a paradigm shift away from the limited static dataset 
setup that is prevalent in much of the work of the community.  

Future work should study the resulting publicly released data to explore other methods of 
lifelong learning, 
or other learning signals that could be extracted from human utterances, for example
 the ideas in \citet{hancock2019selffeeding}.  Another possible direction, for when model performance begins to saturate, is to exploit control of the game engine itself to emphasize learning on the most difficult cases or the ones with the most learning signal, such as in the work on adversarial collection \cite{yang2017mastering,nie2019adversarial}.
 Finally, our role-playing setup can also be applied more generally, for example incorporating both dialogue and actions,  situated in other domains.

\appendix
\section{Game Screenshots}

\begin{figure*}[h!]
\includegraphics[width=16cm]{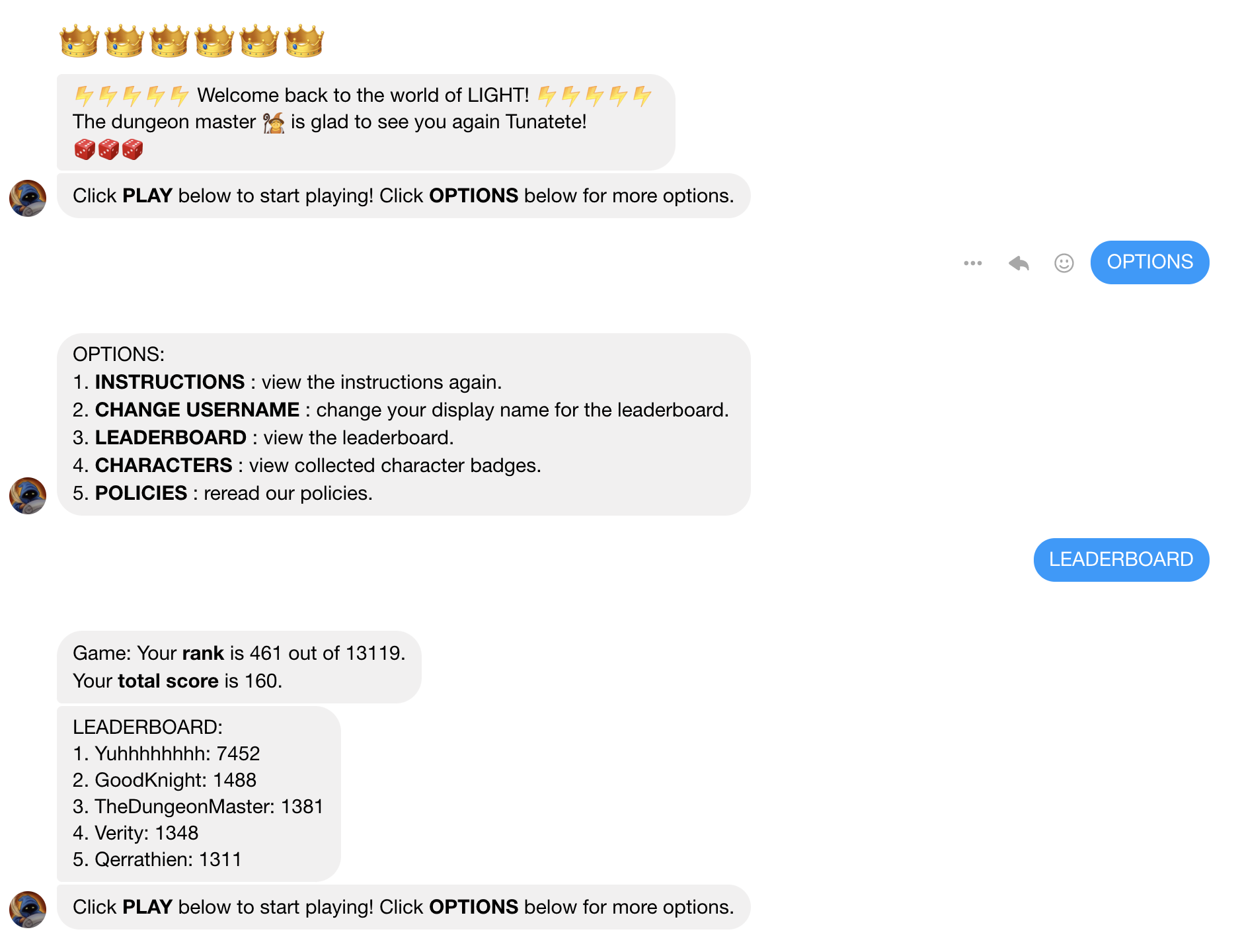}
\includegraphics[width=16cm]{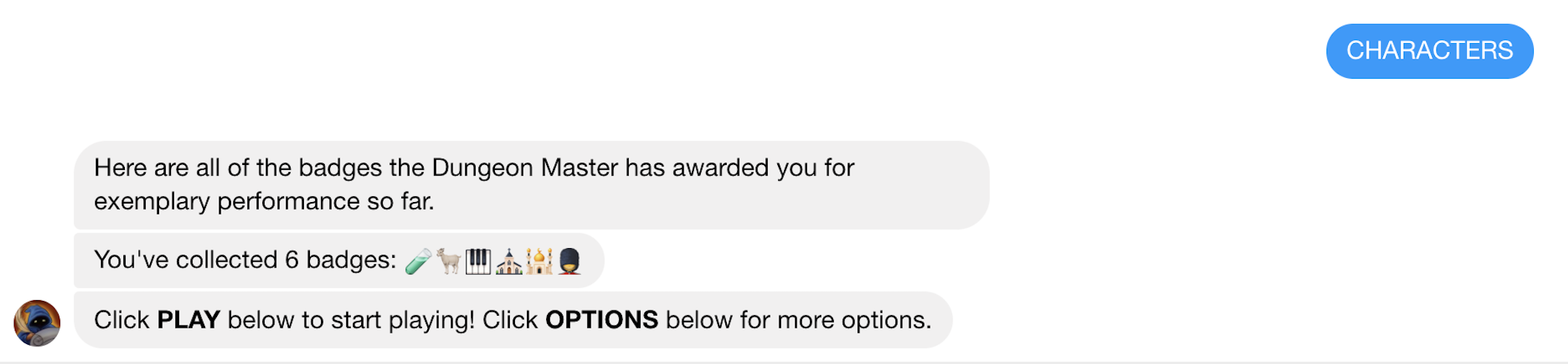}
\includegraphics[width=16cm]{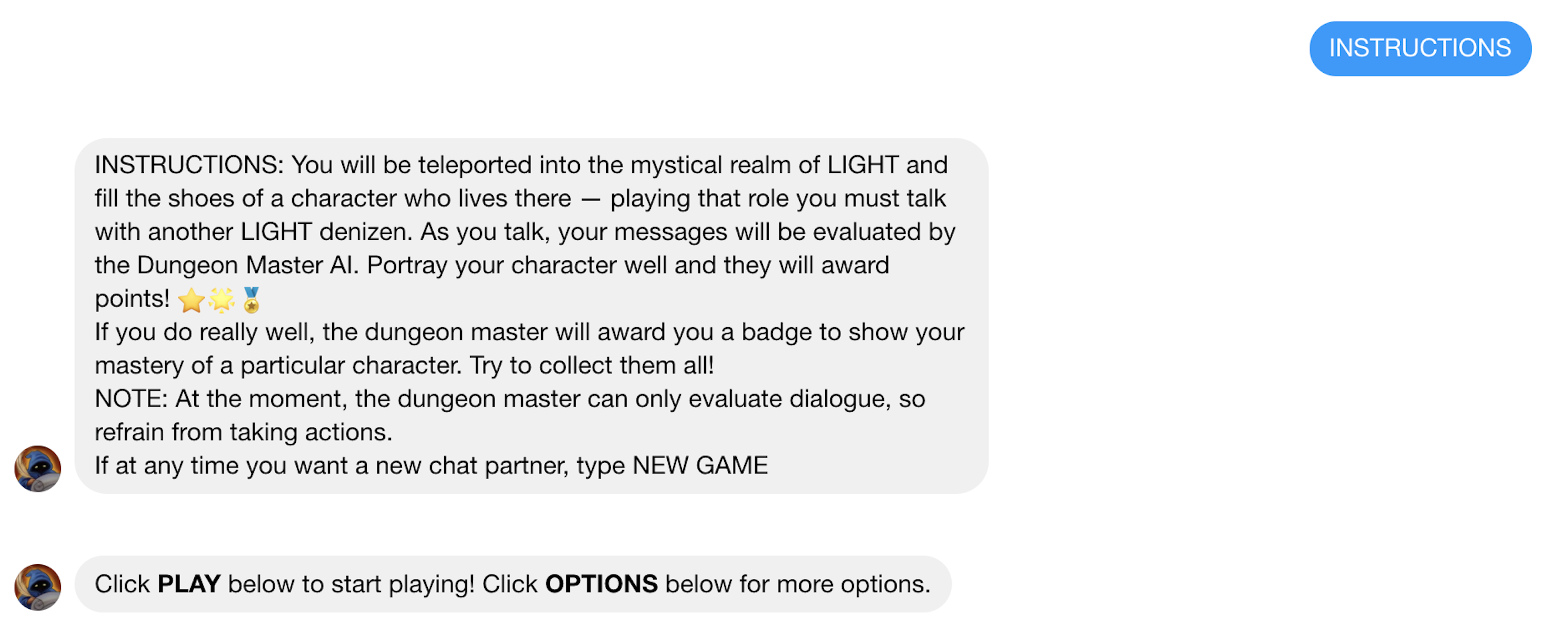}
\vspace{1.5cm}
\caption{Screenshots of the instructions of the LIGHT role-playing game.
\label{fig:game_screenshots}
}
\end{figure*}

\begin{figure*}[h!]
\includegraphics[width=16cm]{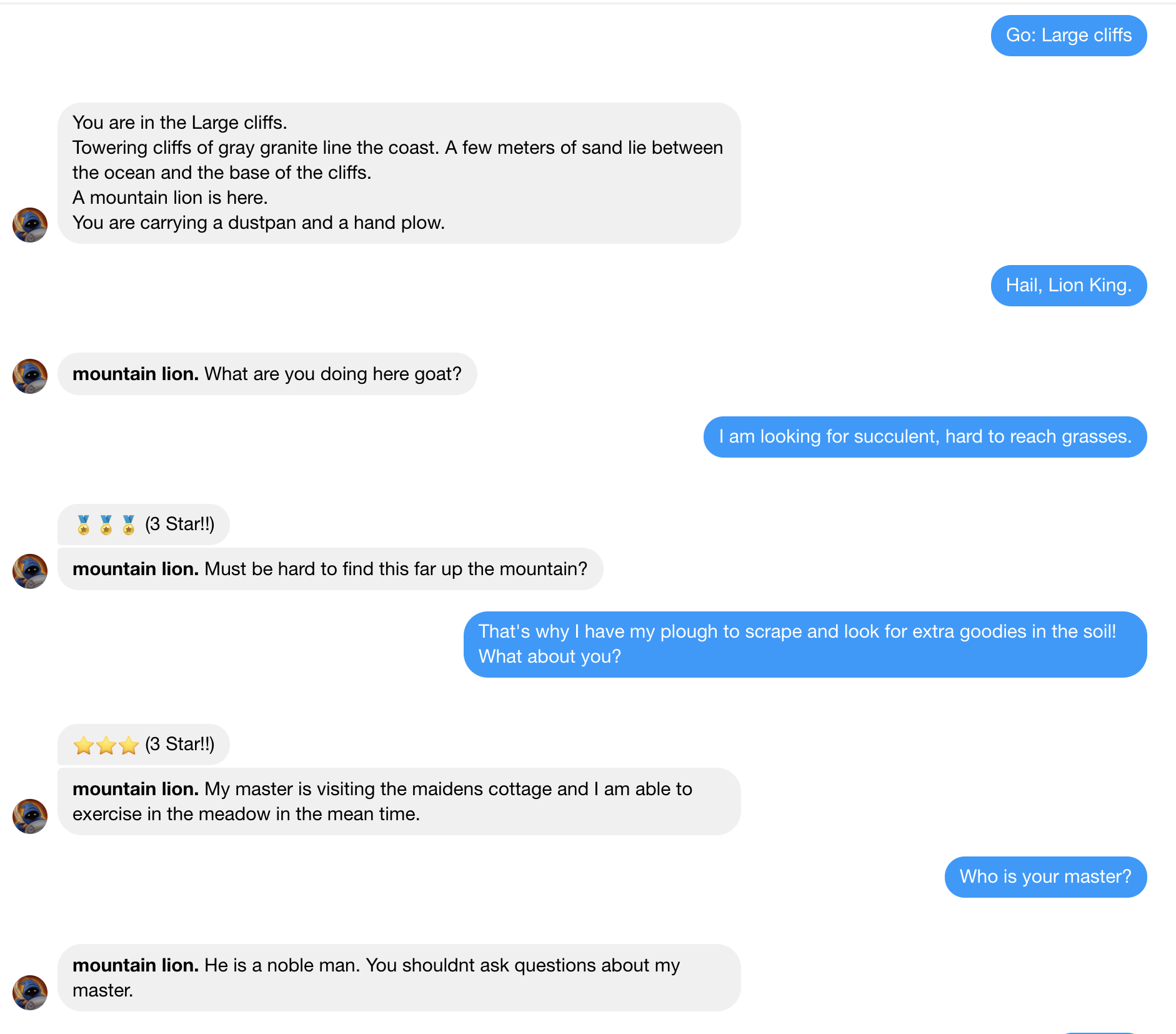}
\\
\\
\includegraphics[width=16cm]{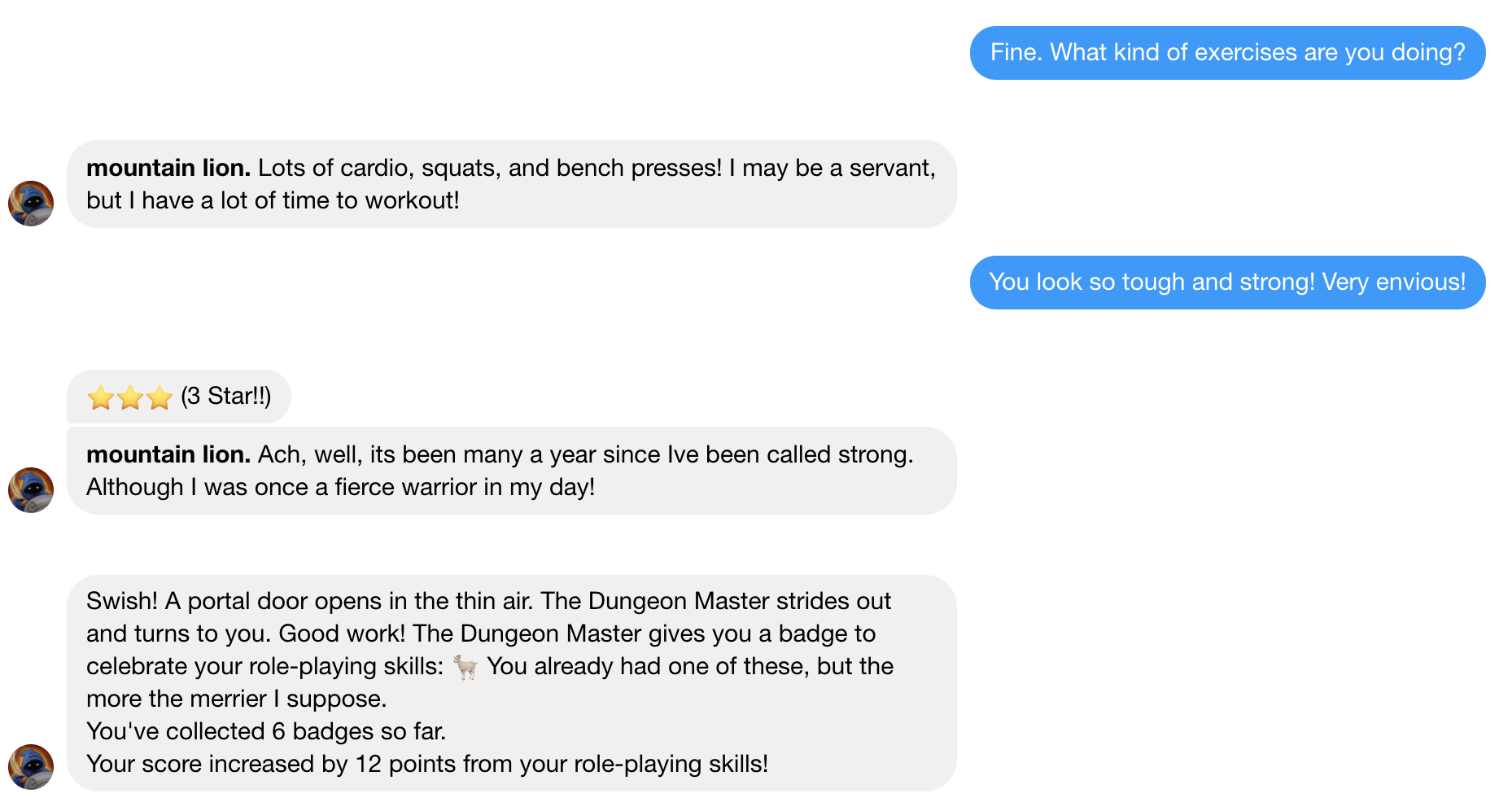}
\vspace{1cm}
\caption{Gameplay Screenshots of the LIGHT role-playing game.
\label{fig:game_screenshots2}
}
\end{figure*}

We show game screenshots in Figures \ref{fig:game_screenshots} and  \ref{fig:game_screenshots2}.

\section{Using WILD Model responses}
In our main results we use the human utterances collected from the role-playing game to form the WILD dataset targets, the hypothesis being that model utterances may or not be correct, and so are not as good a signal. A contrasting view could see training on the model utterance data as a kind of distillation of the previous model's knowledge.
To test the performance of WILD human vs. model utterances, we conducted further experiments comparing them to each other.
We observe a significant drop in Hits@1/20 performance using the model utterances 
for training
 on the WILD validation set (86.05\% for human utterances, and 73.96\% for model),
and similarly on the LIGHT validation set (82.32\% for human utterances, and 77.56\% for model).

\begin{table*}[h]
\center
\begin{small}
\begin{tabular}{l|cccc| c  c}
\toprule
Model  &  Contradiction & Mistaken Identity or Location & Off-topic & Repetitive or Boring & Rating & Count\\ 
\toprule
90M PE 	&  3.9\%  & 9.1\% & 8.4\% & 2.1\% & 3.67 & 88\\			
622M PE &  3.0\%  & 5.3\% & 6.0\% & 5.3\% & 3.69 & 91\\
\bottomrule
\end{tabular}
\end{small}
\caption{Percentage of utterances flagged with an issue alongside overall satisfaction, by model. 
\label{tab:q-function}
}
\end{table*}

\section{Comparing Small and Large Model Variants}
\label{appendix:smallbig}
We evaluated the differences between the small and large retrieval models (Sec. \ref{sec:models}) during deployment to analyze the differences between them. The expectation tends to be that larger models would perform better, especially when the automatic metrics reflected this case. To eliminate the possibility that difficulty in tuning the decoding control differed between the small and large models, we launched a task to crowdworkers to evaluate the models shown in rows 3 and 8 of Table \ref{tab:continuerates}. These correspond to the small and large models trained with LIGHT + WILD data without decoding control, further noted as LW90M and LW622M.

Workers were asked to have a 12-turn conversation with a model (6 turns each), and evaluate each model utterance with respect to some of the most common mistakes we observed models making. They were then asked to provide an overall score for the model, answering "How much fun did you have Role-playing with your chat partner? Would you have a similar conversation with this chat partner in the future?". Results are given in Table \ref{tab:q-function}, listing the percentage of utterances falling into each mistake type.

We observed that LW622M avoided a number of common mistakes when compared to LW90M, contradicting itself less frequently, assuming the role of the wrong character or using a wrong location less frequently, and going off-topic less frequently. These types of mistakes were common complaints of players interacting with our game, and all seemed to be related to having somewhat bad experiences. LW622M also had a higher percentage of conversations without any listed issues, with 34.1\% compared to 27.3\% for LW90M.

LW622M however was rated as being repetitive or boring more than twice as frequently as the smaller model. This difference could be supported in that LW622M used utterances that were on average 2 words shorter than those of LW90M over these evaluations. This issue could explain the phenomenon where players are more likely to leave after interacting with a larger model than a smaller one. The only thing really encouraging unpaid players to continue interacting with our models is that they are fun and engaging, and while it is possible to overlook the model making a mistake if it is still somewhat fun, it is likely less possible to remain engaged when the model is actually boring and repetitive.

While LW622M is a better model across several aspects, it is clear from our live deployment evaluations that something is lost in scaling up from LW90M. Comparing these models with real players lets us see this issue, and moving forward should lead us to search for a model that does not become more boring when learning to not make mistakes.



\bibliography{our}
\bibliographystyle{acl_natbib}

\end{document}